%% file: main.tex
\documentclass[10pt,twocolumn,letterpaper]{article}

\newif\ifARXIV
\ARXIVtrue

\newif\ifCAMERA
\CAMERAfalse

\ifARXIV
    \ifCAMERA
    \usepackage{wacv}
    \else
    \usepackage[pagenumbers,algorithms]{wacv}
    \fi
\else
    \ifCAMERA
    \usepackage{wacv}
    \else
    \usepackage[review,algorithms]{wacv}      
    \fi
\fi

\usepackage{graphicx}
\usepackage{amsmath}
\usepackage{amssymb}
\usepackage{booktabs}
\usepackage{enumitem}

\usepackage[pagebackref,breaklinks,colorlinks]{hyperref}

\usepackage[capitalize]{cleveref}
\crefname{section}{Sec.}{Secs.}
\Crefname{section}{Section}{Sections}
\Crefname{table}{Table}{Tables}
\crefname{table}{Tab.}{Tabs.}

\def\wacvPaperID{2448} 
\def\confName{WACV}
\def\confYear{2025}

\newif\ifDEBUG
\DEBUGfalse

\newif\ifSUPPLEMENTAL
\SUPPLEMENTALtrue

\usepackage[table,xcdraw,dvipsnames]{xcolor}      

\ifDEBUG
    \newcommand{\NJE}[1]{\textcolor{red}{[NJE: #1]}}
    \newcommand{\PJ}[1]{\textcolor{blue}{[PJ: #1]}}
    \newcommand{\JD}[1]{\textcolor{purple}{[JD: #1]}}
    \newcommand{\YHL}[1]{\textcolor{olive}{[YHL: #1]}}
    \newcommand{\GKT}[1]{\textcolor{violet}{[GKT: #1]}}
    \newcommand{\GL}[1]{\textcolor{red}{[GL: #1]}}
    \newcommand{\BC}[1]{\textcolor{cyan}{[BC: #1]}}
\else
    \newcommand{\NJE}[1]{}
    \newcommand{\PJ}[1]{}
    \newcommand{\JD}[1]{}
    \newcommand{\YHL}[1]{}
    \newcommand{\GKT}[1]{}
    \newcommand{\GL}[1]{}
    \newcommand{\GR}[1]{}
    \newcommand{\BC}[1]{}
\fi

\usepackage{multirow}

\newcommand{\blueuparrow}{\textcolor{blue}{$\uparrow$}}
\newcommand{\bluedownarrow}{\textcolor{blue}{$\downarrow$}}

\begin{document}

\title{Token Turing Machines are Efficient Vision Models}


\author{Purvish Jajal\\
Purdue University\\
\and
Nick John Eliopoulos\\
Purdue University\\
\and
Benjamin Shiue-Hal Chou\\
Purdue University\\
\and
George K. Thiruvathukal\\
Loyola University Chicago\\
\and
James C. Davis\\
Purdue University\\
\and
Yung-Hsiang Lu\\
Purdue University\\
}


\maketitle

\begin{abstract}
We propose Vision Token Turing Machines (ViTTM), an efficient, low-latency, memory-augmented Vision Transformer (ViT).
Our approach builds on Neural Turing Machines (NTM) and Token Turing Machines (TTM), which were applied to NLP and sequential visual understanding tasks.
ViTTMs are designed for non-sequential computer vision tasks such as image classification and segmentation.
Our model creates two sets of tokens: \textit{process} tokens and \textit{memory} tokens; process tokens pass through encoder blocks and read-write from memory tokens at each encoder block in the network, allowing them to store and retrieve information from memory.
By ensuring that there are fewer process tokens,
we are able to reduce the inference time of the network while maintaining its accuracy.
On ImageNet-1K, the state-of-the-art ViT-B has median latency of 529.5ms and 81.0\% accuracy, while our ViTTM-B is 56\% faster (234.1ms), with 2.4$\times$ fewer FLOPs, with an accuracy of  82.9\%. 
On ADE20K semantic segmentation, ViT-B achieves 45.65mIoU at 13.8 frame-per-second (FPS) whereas our ViTTM-B model acheives a 45.17 mIoU with 26.8 FPS (+94\%).
\end{abstract}

\section{Introduction}
Vision Transformers (ViTs)~\cite{dosovitskiy_image_2020} are used for a variety of vision tasks such as image recognition, semantic segmentation, object detection, and even image generation.
ViTs are a building block for many applications and models, as both backbones and foundation models. 
However, high-performing ViTs incur large computational costs due to their quadratic complexity with respect to input size. 
Computational costs can be reduced by processing fewer tokens. 
Methods such as token sparsification remove uninformative tokens during inference~\cite{dyvit_2021, tomebolya2023token}.
However, sparsification compromises accuracy and fine-tuning or architectural modifications are required to recover accuracy.

\begin{figure}[t]
    \centering
    \includegraphics[width=0.90\linewidth]{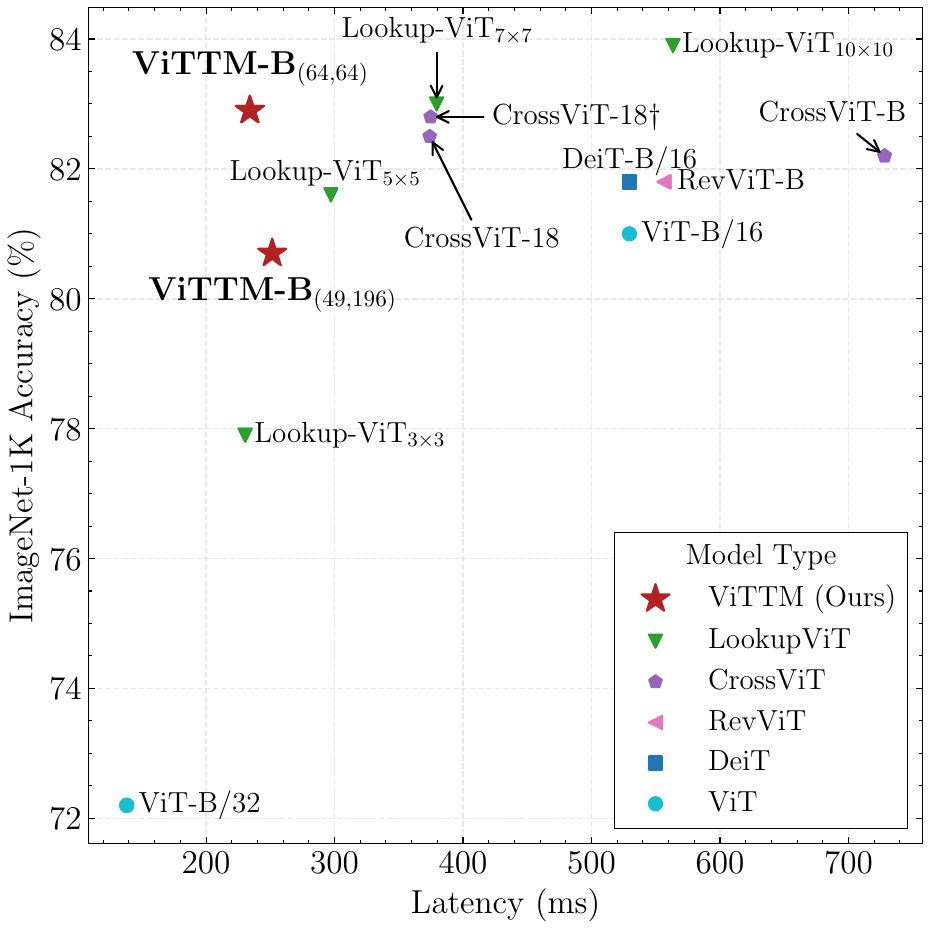}
    \caption{
    Comparison of our architecture with state-of-the-art methods.
    It is evident that ViTTM-B$_{(64,64)}$ has lower latency than its accuracy equivalents (\eg Lookup-ViT$_{7\times7}$) while having higher accuracy than its latency equivalents (\eg Lookup-ViT$_{3\times3}$).
    }
    \label{fig:capstone_pareto}
\end{figure}

Architectural components such as memory mechanisms can improve accuracy in various tasks~\cite{graves_neural_turing_2014, memory_networks_2015, scaling_memory_augmented_2016}.
They are fixed size, and prior work has effectively used memory to enhance models on sequential language and algorithmic tasks~\cite{graves_neural_turing_2014}.
ViTs trained with memory can significantly improve their accuracy~\cite{sandler2022fine}.
Recently, Token Turing Machines (TTMs)~\cite{token_turing_2023} have employed memory with Video Transformers to efficiently process video.
LookupViT~\cite{lookup_vit_2024} uses lookup tokens (akin to memory) to efficiently process images.

\begin{figure*}[t!]
    \centering
    \includegraphics[width=\linewidth]{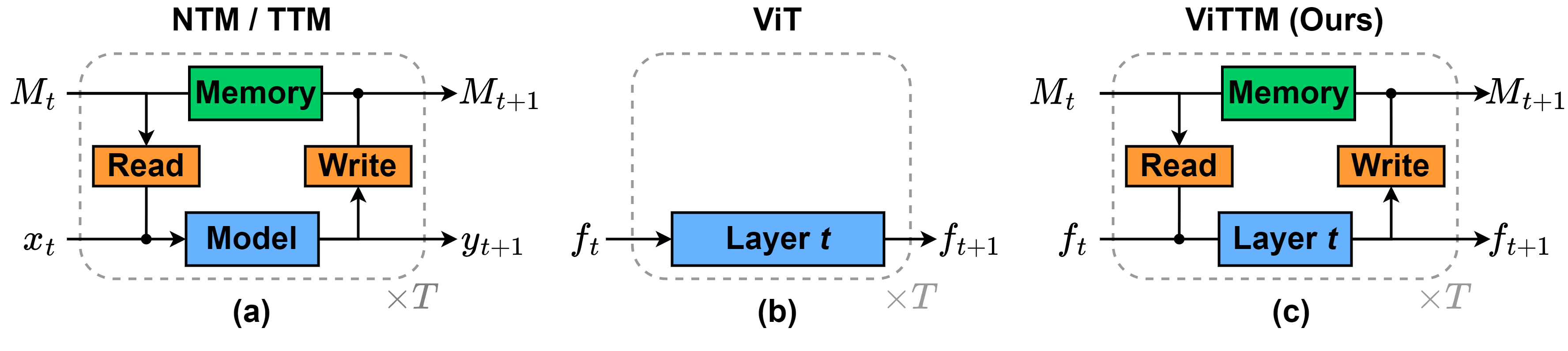}
    \caption{
    Comparison of NTM/TTMs, ViTs, and ViTTMs.
    \textit{(a)} NTMs are sequential models that process an input sequence of size $T$, where inputs $x_t$ are processed at each time step $t$ and the memory $M$ is read from and written at each time step.
    \textit{(b)} ViTs process a single input $f_0$ ($=x_0$), through a series of $T$ layers, where each layer is indexed by $t$, the output features of each layer are denoted $f_t$.
    Our ViTTMs are a synthesis of the NTM and ViT architectures.
    ViTTMs integrate memory into the ViT architecture on a per-layer basis, processing a sequence of features $f_t$ rather than input sequences.
    }
    \label{fig:vittm_ttm_comp}
\end{figure*}

In this work, we present ViTTM, a NTM-ViT hybrid that integrates memory within ViT on a per-layer basis, creating an architecture that is efficient (low latency) and accurate (\cref{fig:vittm_ttm_comp}).
Unlike NTMs, ViTTM processes a sequence of features, not input sequences.
By employing a fixed size memory, ViTTM reduces the number of tokens processed through each layer compared to ViTs.
Integrating memory ensures that ViTTM is \textit{efficient} by processing fewer tokens and \textit{accurate} by storing pertinent information in memory.


We evaluate ViTTM on the ImageNet-1K dataset and the ADE20K semantic segmentation dataset.
ViTTM has competitive accuracy-latency tradeoffs on ImageNet1K classification and expands the Pareto front.
Our ViTTM-B, comparable with ViT-B, requires 2.4$\times$ fewer FLOPs, reducing latency by 56\% 
while having a 1.9\% higher accuracy on ImageNet-1K (82.9\%) than ViT-B.
Our model is also competitive on ADE20K semantic segmentation.
Compared with ViT-B, ViTTM-B achieves an mIoU of 45.17 ($-$0.48) while achieving 94\% higher FPS on an NVIDIA A30 GPU.

We make the following contributions:
\begin{itemize}[noitemsep, topsep=0pt]
    \item We present the ViTTM architecture, an efficient memory-augmented ViT that has competitive accuracy-latency trade-offs on classification and semantic segmentation (\cref{sec:vittm:architecture}).
    \item We conduct a comprehensive ablation of the ViTTM architecture, exploring various designs and identify key decisions that impact the quality of the architecture (\cref{sec:eval:ablations}).
\end{itemize} 


\section{Related Work}
\label{sec:related}
In this section we present prior work in ViTs, token-sparsification, and memory-augmented neural networks.
\cref{sec:related:vit} presents prior work in ViTs and highlight their differences with ViTTM.
We present prior work in token sparsification in \cref{sec:related:sparsification} and highlight the benefits of ViTTM over sparsification.
Prior work in memory-augmented neural networks (MANNs) is presented in \cref{sec:related:mann} and we highlight how we adapt MANNs to non-sequential tasks.

\subsection{Vision Transformers}
\label{sec:related:vit}
After transformers achieved success in language processing tasks \cite{vaswani_attention_2017}, researchers adapted transformer models for computer vision tasks. 
ViT~\cite{dosovitskiy_image_2020} shows that transformers are effective on computer vision tasks.
Subsequent works have focused on developing: (1) training regimes that improve ViT task performance, and (2) alternative ViT architectures to achieve better accuracy-latency trade-offs. 

First, alternative training recipes and self-supervised training are effective at improving ViTs.
Touvron \etal show that ViTs can achieve high accuracy by using data augmentation strategies, without using large datasets~\cite{touvron_deit_2021, touvron_deit_2022}.
Steiner \etal conduct an in-depth empirical study on the training of ViTs~\cite{how_to_train_your_vit_2022}. 
They highlight the interplay between \textit{compute}, \textit{data}, \textit{model size}, and \textit{regularization} for the effective training of ViTs.
In addition, self-supervised learning (SSL) has become a feature of ViT training~\cite{mae_2022, caron2021emerging, chen2021empirical}.

Second, a variety of ViT architectures have been proposed that offer different, or strictly better, accuracy-latency trade-offs.
These include hierarchical ViTs like SwinTransformer and MViT, which create multi-scale feature maps, and CNN-ViT hybrids that incorporate convolutional biases~\cite{liu2021swin, MVITFan_2021_ICCV}.
Among these are \textit{two-stream} architectures --- our work is most similar to these.
Cross-ViT~\cite{cross_vit_2021} one of the first two-stream ViT designs, processes tokens at different patch sizes in separate branches and then merges them through an efficient cross-attention mechanism. 
ViT-CoMer~\cite{vit_comer_2024} utilizes bi-directional interactions with CNNs using a fusion block.
Reversible Vision Transformers~\cite{reversible_vit_2023} create a two-stream architecture with the goal of decoupling GPU memory requirements from model depth by reducing activation caching during training.
A contemporaneous work, LookupViT~\cite{lookup_vit_2024}, processes inputs using two sets of tokens, higher resolution (lookup) tokens and compressed tokens, with information being exchanged between them. 


In this work we present a novel two-stream architecture. 
Unlike prior two-stream architectures, we conduct the majority of the computations in a lightweight stream, thus being computationally efficient.
We highlight differences between our work and prior work in~\cref{sec:vittm:intuition}.
The closest work, LookupViT is conceptually similar but differs in design.

\subsection{Token Sparsification}
\label{sec:related:sparsification}
Token sparsification is a technique used to accelerate ViTs by reducing the overall number of tokens, which also reduces compute cost.
However, sparsification will degrade accuracy without a good heuristic to distinguish informative and uninformative tokens~\cite{intriguing_vit_2021}.
Sparsification methods can be broadly categorized into two approaches: \textit{pruning} methods and \textit{merging} methods~\cite{Haurum_2023_ICCV}.

Pruning-based methods identify and remove uninformative tokens during inference, thereby reducing computation costs.
Popular approaches include using prediction modules~\cite{dyvit_2021}, or using intermediate attention computations as a heuristic to identify unimportant tokens~\cite{fayyaz2022adaptive, yin2022vit, kim2022learned, evitliang2022patches}.

Merging-based methods reduce the total token count by identifying and combining similar tokens.
Boyla \etal propose a training-free merging approach, which averages similar tokens at every layer~\cite{tomebolya2023token}.
Renggli \etal uses learned modules to merge tokens at a single layer, reducing tokens of up to 24$\times$ and consequently large reductions in computational costs~\cite{renggli2022learning}.
Similary, Ryo \etal improve ViTs on video tasks by learning to merge tokens~\cite{tokenlearner_2022}.

Similar to token sparsification, our ViTTM architecture processes fewer tokens at a time by increasing patch size though our approach is not necessarily analogous to pruning or merging methods.
Unlike sparsification, we have access to a memory unit that mitigates the accuracy degradation associated with removing tokens.

\subsection{Memory Augmented Neural Networks}
\label{sec:related:mann}
Memory Augmented Neural Networks (MANNs) are an extension to neural networks that enhance their capabilities with memory~\cite{scaling_memory_augmented_2016}. 
Originally, MANNs enabled RNNs to better tackle various sequential algorithmic, language, and reasoning tasks.
Subsequently, memory has been used to enhance neural networks in various vision tasks.

Neural Turing Machines are one of the first MANNs~\cite{graves_neural_turing_2014}. 
NTMs consist of an RNN equipped with a set of read-write heads that interact with an external memory bank, these modifications allow NTMs to outperform RNNs in copy, sorting, and algorithmic tasks.
Memory Networks are a concurrent approach that focuses upon language and reasoning tasks~\cite{memory_networks_2015}.
Differentiable Neural Computers (DNCs) further extend these NTMs by adding dynamic memory allocation~\cite{graves2016hybrid}. 
Similar to prior works, DNCs show effectiveness on a variety of algorithmic, language, and reasoning tasks.
Recent works, augment transformers with memory sequence modelling~\cite{wu-etal-2022-memformer}.

Recent work has adapted MANNs to both sequential and non-sequential vision tasks.
ViTs have been augmented with learnable memory tokens at every layer to enhance fine-tuning~\cite{sandler2022fine}.
Token Turing Machines (TTM) extend NTMs to transformer models for sequential visual understanding tasks, such as video activity detection~\cite{token_turing_2023}. 
TTM uses an external memory for reading and writing, and introduce a token summarization module that maintains constant computational cost regardless of sequence length.
Our insight is that NTMs can be extended to ViTs by treating intermediate features as a sequence.
In contrast to prior work, we create an architecture that uses memory for the processing of non-sequential tasks.

\section{Vision Token Turing Machines (ViTTM)}
We begin by providing the intuition behind ViTTMs and how this work is related to prior work and draw parallels with NTMs in \cref{sec:vittm:intuition}.
Then, we elaborate on the ViTTM architecture in \cref{sec:vittm:architecture}. 

\begin{figure}[t]
    \centering
    \includegraphics[width=0.57\columnwidth]{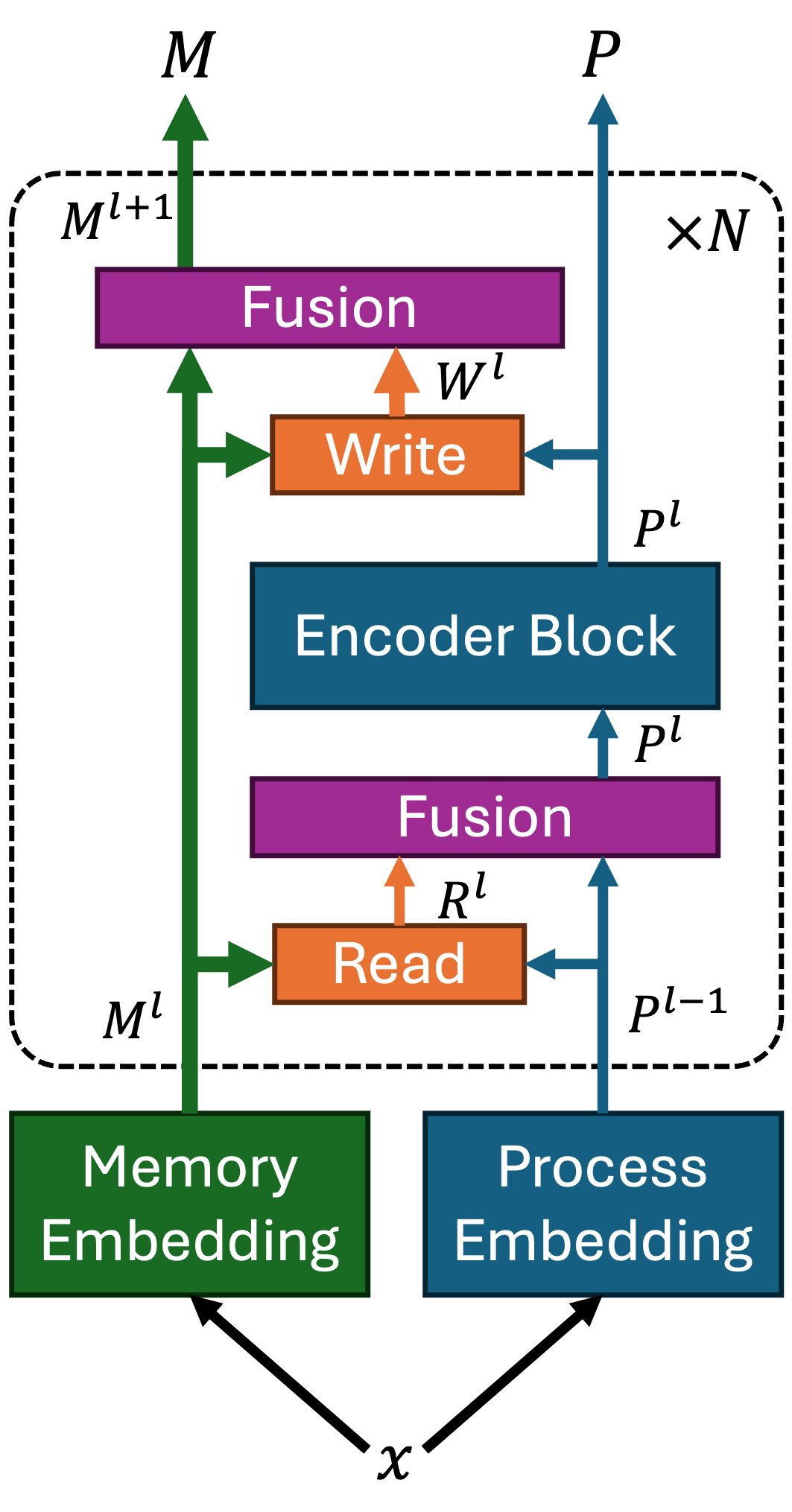}
    \caption{
    ViTTM Architecture. 
    The ViTTM architecture is a NTM-ViT hybrid. 
    In particular, ViTTM  creates two views (or streams) of an input image, $x$, using two patch embedding layers.
    The memory stream, $M$, is created by a memory embedding layer, whereas the process stream, $P$, is created with a process embedding layer. 
    Choose the memory stream to contain a greater number of tokens than the process \ie $T > K$. 
    The process and memory streams exchange information  using \textit{read} and \textit{write} layers, followed by a fusion operation. 
    }
    \label{fig:vittm_arch}
\end{figure}

\subsection{Intuition and Comparison to Prior Work}
\label{sec:vittm:intuition}
The ViTTM architecture is inspired by two key insights:
(1) NTMs and TTMs have shown that neural networks can learn to use memory~\cite{graves2016hybrid, token_turing_2023};
(2) \textit{fewer tokens} through transformer blocks makes inference \textit{faster}, and \textit{more tokens} mean \textit{better accuracy}~\cite{intriguing_vit_2021}.
By integrating these insights, ViTTMs are able to process only a few tokens at a time (\textit{efficiency}), while being able to access the information stored in many tokens (\textit{accuracy}) by using memory.
The end result is an architecture that is both efficient and accurate.

ViTTMs processes images using two token streams, one stream is compute-heavy but processes fewer tokens, while the other contains a larger number of tokens.
Our intuition is that we can learn to exchange information between the two streams using \textit{read-write heads}. 
The information in the compute-heavy stream undergoes more processing, thus we call them \textit{process} tokens. 
Likewise, following NTM, we expect the stream with more tokens to be used to recall and store information, so we call them \textit{memory} tokens. 

This approach distinguishes ViTTMs from previous two-stream architectures like CrossViT and ViT-CoMer, as ViTTM concentrates substantial processing in a single stream.
And, unlike Reversible ViTs, which reduce activation caching during training, ViTTM is specifically designed for computational efficiency during inference.
Furthermore, ViTTMs offer advantages over token sparsification techniques.
By processing fewer tokens while storing their information in memory, ViTTMs maintain both accuracy and efficiency. 
Unlike NTMs, which leverage memory for sequential tasks, our work applies memory to ViTs for a non-sequential task, image recognition. 
 \cref{fig:vittm_ttm_comp} shows a comparison of NTMs to ViTTMs.

\textbf{Comparison to LookupViT (ECCV 2024):}
The contemporaneous LookupViT~\cite{lookup_vit_2024}, is also guided the observation that \textit{processing fewer tokens through transformer blocks makes inference faster}.
Similar to our approach, they exploit this fact by processing two sets of tokens, with the fewer tokens being relegated to the compute-heavy stream.

However, our effort differs from LookupViT, both conceptually and in implementation.
LookupViT primarily utilizes cross attention, while we study various read-write and fusion mechanisms (\cref{sec:eval:ablations}) following prior work in NTMs.
Furthermore, in contrast with LookupViT, we do not refine memory tokens.
This follows NTM design, where memory is solely \emph{read from} or \emph{written to} at any time.
Our decisions 
expand the LookupViT Pareto front by creating a model that is faster while being more accurate (\cref{fig:capstone_pareto}).

\subsection{ViTTM Architecture} \label{sec:vittm:architecture}
 \cref{fig:vittm_arch} provides an overview of the ViTTM architecture.
The ViTTM architecture is synthesis of the ViT and NTM architectures.
Like ViT we convert an input image, $x \in \mathbb{R}^{3 \times H \times W}$, into non-overlapping patches.
But unlike ViT, we create two sets of patches, not one, and convert each set into tokens.
These patches are converted using embedding blocks into $T$ \textit{memory} tokens and $K$ \textit{process} tokens of dimension $d$.

We define the memory and process tokens (streams) as $M\in\mathbb{R}^{T\times d}$, and  $P\in\mathbb{R}^{K\times d}$, respectively.
After embedding, the tokens are processed using $N$ repeating blocks.
We index each block with an index $l$.
Each of these ViTTM blocks, performs three operations in sequence: (1) \textit{read from memory}, (2) \textit{processing}, and (3) \textit{write to memory}.
The read and write operations return ``read tokens" and ``write tokens".
We define the read and write tokens as $R \in \mathbb{R}^{K\times d}$ and $W \in \mathbb{R}^{T\times d}$, respectively.
To ensure efficiency, we depart from ViT's approach of processing all tokens and instead apply the \textit{processing} step only to the process tokens.
The sequence is formally defined as:
\begin{align}
    R^{l} &= \mathrm{Read}(P^{l-1}, M^{l}) \label{eq:read}\\
    P^{l} &= \mathrm{Encoder}(\mathrm{Fusion}(P^{l-1}, R^{l})) \label{eq:encoder}\\
    W^{l} &= \mathrm{Write}(P^{l}, M^{l})\label{write}\\
    M^{l+1} &= \mathrm{Fusion}(M^{l}, W^{l}) \label{eq:write_fusion}
\end{align}

The read and write operations  are performed using \textit{read-write} heads. 
Prior to the \textit{processing} operation the read tokens are merged into the process tokens using a \textit{fusion} operation, and a processing operation is carried out by the encoder block.
After the \textit{processing} operation the write tokens are merged into memory tokens.


The goal of ViTTMs is to be efficient while preserving accuracy.
From \cref{fig:vittm_arch} it is evident that the number of process tokens and the efficiency of the read-write heads will influence latency. 
This is because the encoder block's latency scales with the number of process tokens (\cref{sec:related:sparsification}), and repeated read/write can negate the benefits of reducing tokens. 
In addition, the effective use of memory impacts the task performance. 
Thus, for ViTTMs to be both efficient and effectively utilize memory, four core design decisions must be addressed: 
\begin{enumerate}
  \setlength{\itemsep}{1.5pt}
  \setlength{\parskip}{1.5pt}
  \setlength{\parsep}{1.5pt}
    \item How to initialize memory and process tokens?
    \item How should tokens be selected for read/write?
    \item How will the selected tokens be fused into the process/memory tokens?
    \item Should memory tokens undergo processing?  
\end{enumerate}

Next we present the most effective choices for these design decisions, as identified by our ablations (\cref{sec:eval:ablations}).



\subsubsection{Initializing Process and Memory Tokens} \label{sec:vittm:architecture:process-memory}
Our tokenization process for memory tokens follows the original ViT architecture~\cite{dosovitskiy_image_2020}. 
The process is two steps:
First, we convert an image, $x \in \mathbb{R}^{3 \times H \times W}$ into a set of non-overlapping patches of size $p$.
Second, we apply a learned linear projection, $f: \mathbb{R}^{3 \times H \times W} \mapsto \mathbb{R}^{\frac{H}{p} \times \frac{W}{p} \times d}$, to tokens of length $d$.
The intuition is that the memory tokens should contain a rich representation of the image.
However, the \textit{process} tokens requires further consideration, because it is not obvious how they should be created.

We explore three options for initializing the \textit{process} tokens:
(1) \textit{Latent} -- learning a set of latent tokens; 
(2) \textit{Down-sample} -- down-sampling the memory tokens to create the process tokens; and 
(3) \textit{Patch} -- learning a projection for the process tokens.
In practice, they have identical performance. 
We opt to \textbf{learn separate projections for \textit{memory} and \textit{process} tokens}, since this is most similar ViT embeddings.
To maintain ViTTM's efficiency, we ensure that there are fewer process tokens than memory tokens. 
This is achieved by using a larger patch size for process tokens.


\subsubsection{Read-Write Heads} \label{sec:vittm:architecture:read-write}
Read-write heads play a crucial role in selecting relevant information to read from or write to memory. 
A key consideration is the efficiency of reads and writes. 
To this end, we evaluate several designs for read-write heads: TokenSummary~\cite{token_turing_2023}, Cross Attention~\cite{vaswani_attention_2017}, Latent Attention~\cite{dolga2024latent}, and Linear Attention~\cite{katharopoulos2020transformers}.

In this work, we adopt \textbf{Linear Attention}, as introduced by Katharopoulos \etal~\cite{katharopoulos2020transformers}, to implement our read-write head.
We justify our use of Linear Attention based on theoretical analysis and empirical results from our ablation studies (\cref{sec:eval:ablations}).
It has the following benefits:
\begin{itemize}[noitemsep, topsep=0pt]
    \item The computational complexity of Linear Attention depends on the length of \textit{one} input sequence, whereas Cross Attention depends on the length of \textit{both} sequences.
    \item Linear attention is independent of sequence length, whereas TokenSummary is not. 
    As a result, off-the-shelf token sparsification methods such as ToMe~\cite{tomebolya2023token} can be used along side ViTTM.
    Additionally, we can use training methods such as MAE~\cite{mae_2022}.
    \item We find that \textit{Linear attention} has similar accuracy to cross-attention, while being more efficient.
\end{itemize}

Our read-write head
computes linear attention between two input sequences $X_1 \in \mathbb{R}^{T\times d}$ and $X_2 \in \mathbb{R}^{K\times d}$.
We compute the query, key, and value matrices using the weight matrices $W_q \in \mathbb{R}^{d\times c}$, $W_k \in \mathbb{R}^{d\times c}$, and $W_v \in \mathbb{R}^{d\times d}$, where $d$ represents the \textit{embedding dimension} and $c$ represents the \textit{latent embedding dimension}. 
The query, key, and value matrices are calculated as follows:
\begin{align}
    Q &=  X_2 W_q \in \mathbb{R}^{K \times c}  \label{eq:q}\\
    K &=  X_1 W_k \in \mathbb{R}^{T \times c} \label{eq:k}\\
    V &=  X_1 W_v  \in \mathbb{R}^{T \times d} \label{eq:v}
\end{align}
Following Katharopoulos \etal, we compute the output sequence, $V'$, as follows:
\begin{align}
    V' & = \phi(Q) \, (\phi(K)^{T} \, V)\\
    \phi(x) &= 1  + \text{elu}(x)
\end{align}
During the \textit{read} operation, we set $X_2 = P^{l-1}$, $X_1 = M^l$, and $V' = R^l$.
Conversely, during the \textit{write} operation, we set $X_2 = M^{l}$, $X_1 = P^l$, and $V' = W^l$.
In our evalutation we use $c = d/4$, based on our ablations (\cref{tab:latent_dim_ablation}).


\subsubsection{Process-Memory Fusion}
\begin{figure}[!t]
    \setlength{\belowcaptionskip}{-8pt}
    \centering
    \includegraphics[width=0.80\columnwidth]{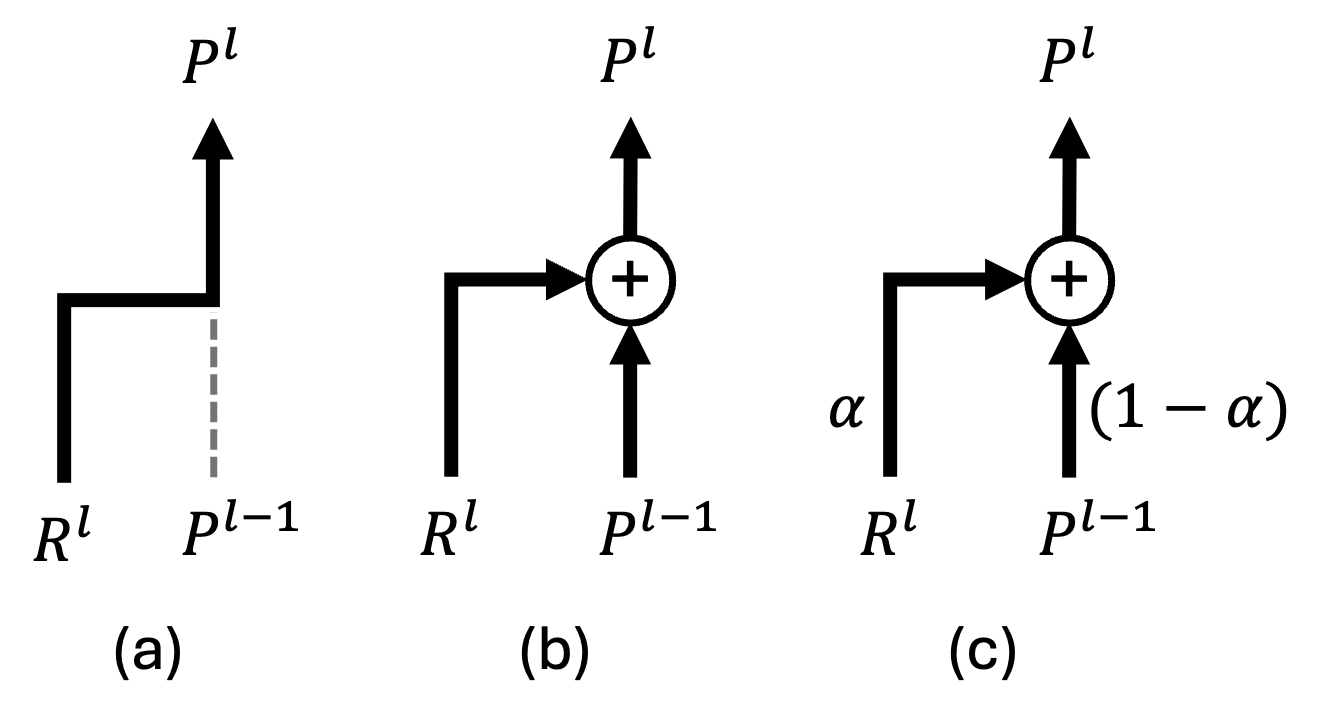}
    \caption{
    Illustration of our fusion implementations. 
    (a) Erase 
    (b) Add 
    (c) Add-Erase. 
    $\alpha$ is computed according to \cref{eq:alpha}.
    Depending on the location of the fusion operation the inputs vary (\cref{fig:vittm_arch}).
    }
    \label{fig:fusion_strats}
\end{figure}

The \textit{fusion} operation combines the read tokens $R^l$ and write tokens $W^l$, from the \textit{read-write heads} into the process and memory streams, respectively.
We explore three implementations of fusion: \textit{erase}, \textit{add}, and \textit{add-erase} fusion (\cref{fig:fusion_strats}). 

\textit{Erase} replaces the previous process tokens ($P^{l-1}$) with the read tokens ($R^l$), i.e. erasing the prior process tokens.
\textit{Add} sums the read tokens ($R^l$) with the process tokens ($P^{l-1}$).
\textit{Add-Erase} creates the new process tokens $P^l$ as a convex combination of the read tokens $R^l$ and the previous process tokens $P^{l-1}$, using combination weights $\alpha$ and $1 - \alpha$.
The combination weights $\alpha$ are computed according to Equation \cref{eq:alpha} (bias omitted).
We project the average read tokens, $R_{avg.} \in \mathbb{R}^d$, using the weight matrix $W_\alpha \in \mathbb{R}^{K \times d}$ and apply the sigmoid non-linearity $\sigma$. 
\begin{align}
    \alpha &= \sigma( W_{\alpha} R^{l}_{avg.}) \in \mathbb{R}^{K}\label{eq:alpha}
\end{align}

In our evaluation we adopt \textbf{Add} fusion because this has the best performance with \textbf{Linear Attention} (\cref{tab:mem_nonlinearity}).

\subsubsection{Processing in the Memory Stream}
\label{sec:vittm:architecture:mem_processing}
Whether the memory tokens should undergo processing is a key consideration. 
Processing allows for the further enrichment of memory tokens, while trading off efficiency.
We explore the effect of adding a multi-layer preceptron (MLP) in the memory stream with varying bottleneck dimensions in~\cref{tab:mem_nonlinearity}.
Unlike LookupViT~\cite{lookup_vit_2024}, 
the benefit of an MLP is marginal and we do not process the memory stream explicitly.
We find that not refining the memory stream can make training unstable --- normalization helps, but increasing batch-size is more effective.

\section{Evaluation}
We evaluate ViTTM design decisions and compare with existing work.
\cref{sec:eval:setup} describes our experimental setup, and \cref{sec:eval:ablations} explores design decisions of ViTTM's.
We compare with other methods on classification and segmentation in \cref{sec:eval:compare_other}.

\subsection{Experimental Setup}
\label{sec:eval:setup}
\textbf{Training:} 
We follow ViT training recipes laid out in~\cite{how_to_train_your_vit_2022}. 
Specifically, we pre-train our models on ImageNet-21K~\cite{imagenet21k}, and subsequently fine-tune them on ImageNet-1K~\cite{imagenet_cvpr09} at 224px image resolution.
For semantic segmentation, we initialize ViTTM backbone models with fine-tuned ImageNet-1K weights. 
We compare against ViT-B/16 backbones fine-tuned at 224px for fair comparison.
Our training recipes are included in the Appendix.

\textbf{Measurement:} FLOP measurements were performed with the \textit{fvcore} library~\cite{fvcore_lib}, and latency was measured using the PyTorch \textit{benchmark} module~\cite{torch_benchmark} with a batch size of 256.
Segmentation experiments and benchmarking were performed using the MMSegmentation library~\cite{mmseg2020}.
Benchmark and evaluations were performed on both a single 80GB A100 GPU and a 24GB A30 GPU.  

\textbf{Notation:} We denote our ViTTM configurations as follows: ViTTM-B$_{(49, 64)}$ is a ``ViTTM-B model with 49 process tokens and 64 memory tokens". 
For models \textit{without} memory tokens (\ie baseline models), we use $\phi$ to denote the absence of memory tokens.

For reproducibility, our code is open source~\cite{efficientttms}. 



\subsection{Ablations}
\label{sec:eval:ablations}
In this section, we perform ablation studies on the four core design decisions of ViTTMs (\cref{sec:vittm:architecture}).
We conduct ablations to answer the following:
\begin{enumerate}
  \setlength{\itemsep}{1.5pt}
  \setlength{\parskip}{1.5pt}
  \setlength{\parsep}{1.5pt}
    \item How to initialize memory and process tokens? (\cref{sec:eval:ablations:initialization} and  \cref{sec:eval:ablations:tokens})
    \item How tokens will be selected for read/write? (\cref{sec:eval:ablations:rw_heads} and \cref{sec:eval:abalations:latent})
    \item How will the selected tokens be fused into the process/memory tokens? (\cref{sec:eval:ablations:rw_heads})
    \item Should memory tokens undergo processing? (\cref{sec:eval:ablations:mem_nonlinearity})
\end{enumerate}
To assess the efficacy of each design choice, we fine-tune and evaluate on ImageNet-1K.
We allocate a fixed amount training time, 100 epochs, for each ablation.
The ablations in \cref{sec:eval:ablations:initialization} and \cref{sec:eval:ablations:rw_heads} are conducted using a ViTTM-S model, whereas those in \cref{sec:eval:abalations:latent}, \cref{sec:eval:ablations:tokens}, and \cref{sec:eval:ablations:mem_nonlinearity} ablations use a ViTTM-B model.

\subsubsection{Memory and Process Initialization}
\label{sec:eval:ablations:initialization}
We evaluate the initialization method for process tokens as defined in~\cref{sec:vittm:architecture:process-memory}.
Ultimately, the choice of initialization scheme has no impact on accuracy.
Therefore, we opted for \textit{Patch} initalization for all of our models since it is the standard method for ViT architectures.


\subsubsection{Read-Write Head and Fusion Choice}
\label{sec:eval:ablations:rw_heads}
We explore combinations of read-write heads and fusion mechanisms outlined in~\cref{sec:vittm:architecture}.
\cref{tab:rw_ablate} displays the ablation results for read-write head and fusion mechanisms.
We observe that all read-write heads provide comparable accuracy, primarily differing in computational cost.
As stated in \cref{sec:vittm:architecture:read-write}, we use Linear Attention~\cite{katharopoulos2020transformers} due to its efficiency and comparable accuracy to Cross Attention.

\input{tables/table_rw_ablate}

\subsubsection{Effect of Read-Write Head Latent Dimension} \label{sec:eval:abalations:latent}
\cref{tab:latent_dim_ablation} illustrates the impact of varying the latent embedding dimension $c$.
Reducing the latent embedding dimension relative to the embedding dimension, $d$, does not reduce accuracy, but can provide substantial latency benefits.
As shown in~\cref{tab:latent_dim_ablation}, setting $c=d/4$ reduces latency by 100ms with negligible impact on Top-1 accuracy.
Given this result, we choose $c=d/4$ for all evaluations in this work.

\input{tables/table_latent_dim_ablation}

\subsubsection{Effect of Memory and Process Token Number}
\label{sec:eval:ablations:tokens}
In \cref{tab:process_mem_ablate}, we present how the number of memory and process tokens affects both the computational cost and accuracy of ViTTM-B.
We find: (1) \textit{memory tokens improve accuracy}, (2) \textit{the marginal accuracy increase of additional memory tokens is small}, and (3) \textit{the marginal accuracy increase of additional process tokens is large}.

\input{tables/table_process_mem_ablate}

First, it is clear that adding memory tokens improves accuracy.
When comparing ViT-B$_{(49, \phi)}$ to ViTTM-B$_{(49, 64)}$, there is a 7.2\% improvement in Top-1 accuracy. 
Second, the marginal benefit of extra memory tokens is minimal.
This is evident when we compare ViTTM-B$_{(49, 64)}$,  ViTTM-B$_{(49, 196)}$, and  ViTTM-B$_{(49, 256)}$, where adding extra memory tokens yield minimal accuracy gains.
Third, the marginal benefit of extra process tokens is high.
As seen when comparing ViTTM-B$_{(49, 64)}$ to ViTTM-B$_{(64,64)}$.
The addition of process tokens, while keeping memory tokens constant, increases accuracy by 1.2\%.

\input{tables/table_mem_nonlinearity}

\subsubsection{Effect of Memory Stream Non-linearity}\label{sec:eval:ablations:mem_nonlinearity}
LookupViT proposed using a non-linearity to their lookup tokens (analogous to our memory stream) for accuracy. 
We report that this is inefficient (\cref{tab:mem_nonlinearity}).
An MLP increases FLOPS by 14\%$-$56\% and latency by 13\%$-$47\%, while having negligible effects on accuracy.

\input{tables/table_vittm_im1k}

\subsection{Comparison with other work}\label{sec:eval:compare_other}
We evaluate our ViTTM models on ImageNet-1K classification (\cref{sec:eval:classification}) and ADE20K semantic segmentation (\cref{sec:eval:seg}) using Segmenter~\cite{strudel2021segmenter} from the MMSegmentation library~\cite{mmseg2020}.
For image classification, we compare against existing two-stream architectures, and use ViT and DeiT models as baselines~\cite{dosovitskiy_image_2020,touvron_deit_2021}.
For semantic segmentation, we compare with a ViT-B/16 model from the timm library~\cite{huggingface_timm}.
Based on our ablations we configure our ViTTM model with \textbf{Linear Attention} read-write heads ($c=d/4$), \textbf{Add} fusion, and with \textbf{no} memory stream non-linearity.

\subsubsection{Image Classification (ImageNet-1K)}\label{sec:eval:classification}
ViTTM offers competitive accuracy-latency trade-offs on ImageNet-1K.
\cref{tab:im1k_eval} presents our evaluation with existing methods.
We achieve lower latency while matching or nearly matching accuracy of other methods, thereby expanding the Pareto frontier (\cref{fig:capstone_pareto}).
For example, compared with the similar latency LookupViT$_{3\times3}$, ViTTM-B$_{(64,64)}$ has 5.5\% greater accuracy.
When compared with the similar accuracy LookupViT$_{7\times7}$, our ViTTM-B$_{(64,64)}$ has a 145.1ms (38\%) lower latency.

\subsubsection{Semantic Segmentation (ADE20K)}
\label{sec:eval:seg}
ViTTM also achieves competitive results on ADE20K across multiple devices, as illustrated in \cref{tab:segmentation}.
Our ViTTM-B$_{(64,64)}$ matches ViT-B/16$_{224}$ on ADE20K semantic segmentation while improving FPS by 37\% (23.8 FPS to 32.5 FPS) on an A100 GPU, and by 94\% (13.8 FPS to 26.8 FPS) on an A30 GPU.
In this work, we do not investigate the reason for the discrepancy between devices. 
\input{tables/table_vittm_ade20k}

\section{Discussion and Future Work}
\textbf{Two-Stream Architectures:} It is evident from both prior work and our experiments that two-stream architectures can be used to create both efficient and accurate vision models, \eg, LookupViT, CrossViT, and ViTTMs.
However, two-stream architectures present more design choices than ViTs, leaving much to explore about what constitutes an optimal design.
For example, although we ablate many design choices (\cref{sec:eval:ablations}), we do not consider differing embedding dimensions per stream, varying depth, or the effect of self-supervised learning.
Previous studies~\cite{cross_vit_2021} have demonstrated the effectiveness of altering embedding dimensions, yet its interplay with other parameters remains unclear.
Changing the embedding dimension has been shown to be effective in prior work~\cite{cross_vit_2021}, but how it is related to other parameters is unknown.
Moreover, we also find that two-stream architectures can suffer from training instabilities (\cref{sec:vittm:architecture:mem_processing}).
An important direction for future work is the rigorous study of both architectural choices and the training dynamics of two-stream architectures. 

\textbf{Read-Write Heads:} An important consideration in our design is the choice of read-write heads. 
As stated in~\cref{sec:vittm:architecture:read-write}, read-write heads must be efficient.
We believe that efficient implementation of read-write mechanisms is an important direction for future work. 
Actually obtaining latency reductions from FLOP reductions is important for the further development and use of alternative architectures.

\textbf{Uses of Memory Stream for Multi-Modal application:} 
In this work we ablate how the process stream is be initialized, \ie how it is populated with tokens (\cref{sec:vittm:architecture:process-memory}).
However, we do not explore how the memory tokens should be initialized. 
It is possible to initialize the memory tokens using another modality, thus creating a multi-modal model. 
Such a model would read/write (fuse) information across modalities.
Recent work, in generative models use two-stream architectures for this purpose~\cite{esser2024scaling, fei2024flux}.

\section{Conclusion}
We introduced ViTTM, a ViT-NTM hybrid, that integrates memory into vision transformers to create an efficient and accurate vision model. 
Evaluation on ImageNet1K and ADE20K demonstrated that ViTTM accuracy matches or exceeds prior work.
ViTTM-B is 56\% faster (2.4$\times$ fewer FLOPs) than ViT-B while having an accuracy of 82.9\% (+1.9\%) on ImageNet1K.
On ADE20K, ViTTM-B achieves 37\%-94\% higher FPS than ViT-B.
Our current work illustrated the effectiveness of ViTTMs, and we anticipate that their efficiency can be improved and extended to other tasks.
Future work that explores latency optimization of the read-write mechanisms, such as fusing operators, would be an effective extension.

{\small
\bibliographystyle{ieee_fullname}
\bibliography{bib/references}
}

\ifSUPPLEMENTAL
\clearpage
\section*{Appendix}
In this appendix, we provide details of our training regime and provide additional evaluation on ImageNet1K.
First, in \cref{appendix:train}, we detail hyperparameters and other details about our training regime for ViTTM.
Second, in \cref{appendix:extra}, we compare ViTTMs with Token Merging \cite{tomebolya2023token}, and show results for a small ViTTM model (ViTTM-S).
Our code can be found at \url{https://github.com/pjjajal/EfficientTTMs}.

\subsection{Training Configurations}\label{appendix:train}
In \cref{sec:eval:setup}, we briefly described our training recipe for ViTTM. 
Here, \cref{tab:app:settings} provides details of both our pre-training and fine-tuning regimes to enhance reproducibility of our work.
"RRC" indicates the use of random resize crop.
The "CE" Loss refers to Cross Entropy, and "BCE" refers to Binary Cross Entropy.

\input{tables/table_appendix_training_parameters}

\subsection{Extra Results}\label{appendix:extra}
\input{tables/table_appendix_extra_1k_results}
We present extra ImageNet-1k results in~\cref{tab:app:extra_1k}.
Specifically, we trained a ViTTM-S model, and include comparisons against ViT-S and ViT-B augmented with Token Merging at various pruning rates (without fine-tunign).
ViTTMs consistently have lower latency than state-of-the-art methods while matching their accuracy.
Compared with Token Merging, ViTTMs achieve higher accuracy (as expected), while having lower latency across a range of pruning ratios ($r$).

\else\fi

\end{document}

%% file: tables/table_rw_ablate.tex
\begin{table}[!b]
    \small
    \centering
    \begin{tabular}{lrrrr}
        \toprule
        \multirow{3}{*}{\textbf{Fusion}} & \multicolumn{4}{c}{\textbf{Read-Write Head}}\\
        ~ & Token & Cross & Latent & Linear\\
        ~ & Summary & Attn. & Attn. & Attn.\\ 
        \toprule
        Erase     &  72.49         & 3.94            &  0.33          &  2.43  \\
        Add       &  75.86         & 76.26           & \textbf{76.17} & \textbf{76.26} \\
        Add-Erase & \textbf{76.26} & \textbf{76.33}  & 75.82          & 74.74   \\
        \midrule
        GFLOPs    & 1.86   & 2.92 & 3.07 & 2.62 \\
        \bottomrule
    \end{tabular}
\caption{
Ablation over read-write head and fusion mechanisms.
Each cell represents the Top-1 ImageNet1K accuracy for a read-write head and fusion mechanism combination.
The GFLOPs are measured for each read-write head using \textit{Add} fusion (fusion contributions to GFLOPs negligible).
TokenSummary has the lowest computational cost, followed by Linear Attention, Cross Attention, and finally Latent Attention.
\textit{Add} fusion works best, whereas \textit{Add-Erase} fusion degrades the performance of Latent and Linear Attention.
\textit{Erase} fusion performs the worst for the attention-based heads, and best when used with Token Summary.
}
\label{tab:rw_ablate}
\end{table}

%% file: tables/table_latent_dim_ablation.tex
\begin{table}[!t]
    \small
    \centering
    \setlength{\belowcaptionskip}{-8pt}
    \setlength{\tabcolsep}{4pt}
    \begin{tabular}{crrr}
        \toprule
        {$c$} & \textbf{GFLOPs} (M)\bluedownarrow & \textbf{Latency} (ms)\bluedownarrow & \textbf{Top-1} (\%)\blueuparrow\\
        \toprule
        $d/4$ & 7.10 &  250.7 &  78.4  \\
        $d/2$& 8.04 &  288.9 &  78.0  \\
        $d$ & 9.92 &  352.2 &  78.3  \\
        \bottomrule
    \end{tabular}
    \caption{
    Ablation over the \textit{latent embedding dimension}, $c$ for the read-write head on ViTTM-B$_{(49,196)}$ ($d=768$).
    Accuracy is computed using the memory tokens rather than the process tokens.
    We use \blueuparrow/\bluedownarrow~notation to indicate \textit{``larger/smaller is better"}.
    We find latency reductions saturate when $c < d/4$.
    }
    \label{tab:latent_dim_ablation}
\end{table}

%% file: tables/table_process_mem_ablate.tex
\begin{table}[!b]
    \small
    \centering
    \setlength{\tabcolsep}{4pt}
    \begin{tabular}{rrrrr}
        \toprule
        \textbf{Process} & \textbf{Memory} & \textbf{GFLOPs}\bluedownarrow & \textbf{Latency} \bluedownarrow & \textbf{Top-1} \blueuparrow\\
        \textbf{Tokens} & \textbf{Tokens} & & (ms) & (\%) \\
        \toprule
        49  & $\phi$ &  4.37  &  138.3 & 72.2\\ 
        196 & $\phi$ &  16.87 &  527.4 & 81.0\\ 
        \midrule
        16 &  64 &   3.39 &  125.0 & 70.4 \\
        16 & 196 &   6.37 &  248.0 & 70.9 \\
        16 & 256 &   7.72 &  298.9 & 71.2 \\
        \midrule
        49 &  64 &   6.94 & 233.9 & 79.4 \\
        49 & 196 &   9.92 & 357.7 & 79.3 \\
        49 & 256 &  11.27 & 408.7 & 79.8 \\
        \midrule
        64 &  64 &  8.56 & 285.3 & 80.6 \\
        64 & 196 & 11.53 & 409.0 & 80.4 \\
        64 & 256 & 12.88 & 460.0 & 80.9 \\
        \bottomrule
    \end{tabular}
    \caption{
    Ablation of number of tokens for the process and memory streams on ViTTM-B; we use the $\phi$ to indicate models \textit{without} memory tokens \ie baseline models.
    Although memory tokens improve accuracy, the marginal improvement is small, \eg, 64 $\rightarrow$ 256 memory tokens. 
    Increasing the number of process tokens has the greatest impact on accuracy, \eg, 49 $\rightarrow$ 64 process tokens.
    }
    \label{tab:process_mem_ablate}
\end{table}

%% file: tables/table_mem_nonlinearity.tex
\begin{table}[b]
    \centering
    \setlength{\tabcolsep}{4pt}
    \small
    \begin{tabular}{lrrrr}
        \toprule
        \textbf{Mem. Stream} & \textbf{Params} & \textbf{GFLOPs} \bluedownarrow & \textbf{Latency} \bluedownarrow & \textbf{Top-1} \blueuparrow\\
        \textbf{Block} & (M) &  & (ms) & (\%) \\
        \toprule
        None          & 37.9 &   9.92 &  358.2 & 78.2\\
        MLP ($r=0.5$) & 39.7 &  11.31 &  403.6 & 78.3\\
        MLP ($r=1.0$) & 41.5 &  12.70 &  444.9 & 78.3\\
        MLP ($r=2.0$) & 45.0 &  15.47 &  525.1 & 78.2\\
        \bottomrule
    \end{tabular}
    \caption{
    Ablation over the choice of non-linearity for the memory stream on ViTTM-B.
    $r$ is the scaling ratio applied to the embedding dimension $d$, thus the hidden dimension of the MLP is $r \times d$.
    The inclusion of an MLP layer seems to have no effect on accuracy while increasing the FLOPs and latency of our model.
    }
    \label{tab:mem_nonlinearity}
\end{table}

%% file: tables/table_vittm_im1k.tex
\begin{table*}[!h]
    \setlength{\tabcolsep}{8pt}
    \small
    \centering
    \begin{tabular}{c|lrrrrr}
        \toprule
        \textbf{Model Class} & \textbf{Model} & \textbf{Params} (M) & \textbf{GFLOPs} \bluedownarrow & \textbf{Latency} (ms)\bluedownarrow & \textbf{Top-1}(\%)\blueuparrow \\
        \toprule
        \multirow{3}{*}{ViT/DeiT} & ViT-B/32~\cite{huggingface_timm} & 88 & 4.37 & 138.3 & 72.2  \\
        & ViT-B/16~\cite{huggingface_timm}           &  87 & 16.87 & 529.5 & 81.0   \\
        & DeiT-B/16~\cite{touvron_deit_2021}          &  87 & 16.87 & 529.7 & 81.8   \\
        \midrule
        \multirow{8}{*}{Two-Stream} & CrossViT-B~\cite{cross_vit_2021}         & 105 & 21.22 & 728.1 & 82.2 \\
        & CrossViT-18~\cite{cross_vit_2021}          & 43  & 9.05  & 374.1 & 82.5  \\
        & CrossViT-18$\dagger$~\cite{cross_vit_2021} & 44  & 9.50  & 378.2 & 82.8  \\
        & Rev-ViT-B~\cite{reversible_vit_2023}        & 86  & 17.49 & 556.5 & 81.8  \\
        & LookupViT$_{3\times3}$~\cite{lookup_vit_2024}     &  90 &   5.26 &  230.5 & 77.9 \\
        & LookupViT$_{5\times5}$~\cite{lookup_vit_2024}     &  90 &   6.94 &  297.2 & 81.6 \\
        & LookupViT$_{7\times7}$~\cite{lookup_vit_2024}     &  90 &   9.45 &  379.5 & 83.0 \\
        & LookupViT$_{10\times10}$~\cite{lookup_vit_2024}   &  90 &  14.80 &  563.4 & 83.9 \\
        \midrule
        \multirow{2}{*}{Ours} & ViTTM-B$_{(64,64)}$    & 127 &  7.08 & 234.1 & 82.9  \\
        & ViTTM-B$_{(49,196)}$ & 125  &  7.10 &  251.5 & 80.9 \\
        
        \bottomrule
    \end{tabular}
    \caption{
    Comparison of ViTTM with state-of-the-art methods on image classification (ImageNet-1K).
    ViTTMs are much faster (lower latency) than  state-of-the-art methods while matching their accuracy is most cases.
    \cref{fig:capstone_pareto} depicts the data in this table.
    Latency was measured on a 80GB A100 with batch size 256.
    \textit{Notes}:
    The ViT baseline model is the 224 resolution fine-tuned model from~\cite{dosovitskiy_image_2020}, available from \texttt{timm}~\cite{huggingface_timm}.
    LookupViT does not have a public implementation, as such we implement a version following the paper.
    }
    \label{tab:im1k_eval}
\end{table*}

%% file: tables/table_vittm_ade20k.tex
\begin{table}[!t]
    \centering
    \small
    \begin{tabular}{lrrr}
        \toprule
        \textbf{Model} (M) 
        & \textbf{FPS (A100)} \blueuparrow 
        & \textbf{FPS (A30)} \blueuparrow
        & \textbf{mIoU} \blueuparrow \\
        \toprule
        ViT-B/16$_{384}$~\cite{strudel2021segmenter} & 23.8 & 13.8 & 48.06 \\
        ViT-B/16$_{224}$ & 23.8 & 13.8 & 45.65 \\
        ViTTM-B$_{(64,64)}$  & \textit{(+37\%)} 32.5 & \textit{(+94\%)} 26.8 &  45.17\\
        ViTTM-B$_{(49,196)}$   & \textit{(+38\%)} 32.8 & \textit{(+94\%)} 26.7 &  43.60 \\
        \bottomrule
    \end{tabular}
    \caption{
    Results for semantic segmentation (ADE20K) using the Segmenter~\cite{strudel2021segmenter} segmentation method.
    FPS measurements are taken on 80GB NVIDIA A100 and 24GB A30 GPUs.
    }
    \label{tab:segmentation}
\end{table}

%% file: tables/table_appendix_training_parameters.tex
\begin{table}[ht]
    \centering
    \begin{tabular}{l l l}
        \toprule
        \textbf{} & \textbf{Pre-training} & \textbf{Fine-Tuning} \\
        \midrule
        Eff. Batch size & 4096 & 2048 \\
        Optimizer & AdamW & AdamW \\
        LR & $1.5\times10^{-4}$ & $0.25\times10^{-4}$ \\
        Warmup LR & $1.0\times10^{-6}$ & $1.0\times10^{-6}$ \\
        Min. LR & $1.0\times10^{-7}$ & $1.0\times10^{-7}$ \\
        LR decay & cosine & cosine \\
        Weight decay & 0.03 & 0.1 \\
        Warmup epochs & 3 & 5 \\
        Epochs & 300 & 300 \\
        \midrule
        Stoch. Depth & $0.1$ & $0.1$ \\
        Gradient Clip. & $1.0$ & $1.0$ \\
        \midrule
        Image Size & 224 & 224 \\
        Horiz. flip & $\checkmark$ & $\checkmark$ \\
        RRC & $\checkmark$ & $\checkmark$ \\
        RandAug Ops & $\times$ & $2$ \\
        RandAug Mag. & $\times$ & $20$ \\
        Mixup alpha & $\times$ & $0.8$ \\
        CutMix alpha & $\times$ & $0.8$ \\
        Erasing prob. & $\times$ & $0.25$ \\
        \midrule
        Loss & CE & BCE \\
        \bottomrule
    \end{tabular}
    \caption{Training configurations for ViTTM-B models. 
    All training was performed on NVIDIA A100 80GB GPU's. 
    }
    \label{tab:app:settings}
\end{table}

%% file: tables/table_appendix_extra_1k_results.tex
\begin{table*}[!ht]
    \setlength{\tabcolsep}{8pt}
    \centering
    \begin{tabular}{c|lrrrrr}
        \toprule
        \textbf{Model Class} & \textbf{Model} & \textbf{Params} (M) & \textbf{GFLOPs} \bluedownarrow & \textbf{Latency} (ms)\bluedownarrow & \textbf{Top-1}(\%)\blueuparrow \\
        \toprule
        \multirow{5}{*}{ViT/DeiT} & ViT-S/16 & 22 & 4.25 & 149.5 & 74.2 \\
        & DeiT-S/16 & 22 & 4.25 & 152.0 & 79.8  \\
        & ViT-B/32 & 88 & 4.37 & 138.3 & 72.2  \\
        & ViT-B/16           &  87 & 16.87 & 529.5 & 81.0   \\
        & DeiT-B/16           &  87 & 16.87 & 529.7 & 81.8   \\
        \midrule
        \multirow{11}{*}{Two-Stream} & CrossViT-S         & 27 & 5.63 & 235.7 & 81.0 \\
        & CrossViT-15          & 28  & 5.81  & 249.1 & 82.3  \\
        & CrossViT-15$\dagger$ & 28  & 6.13  & 252.3 & 81.5  \\
        & CrossViT-B         & 105 & 21.22 & 728.1 & 82.2 \\
        & CrossViT-18          & 43  & 9.05  & 374.1 & 82.5  \\
        & CrossViT-18$\dagger$ & 44  & 9.50  & 378.2 & 82.8  \\
        & Rev-ViT-B        & 86  & 17.49 & 556.5 & 81.8  \\
        & LookupViT$_{3\times3}$     &  90 &   5.26 &  230.5 & 77.9 \\
        & LookupViT$_{5\times5}$     &  90 &   6.94 &  297.2 & 81.6 \\
        & LookupViT$_{7\times7}$     &  90 &   9.45 &  379.5 & 83.0 \\
        & LookupViT$_{10\times10}$   &  90 &  14.80 &  563.4 & 83.9 \\
        \midrule
        \multirow{14}{*}{Token Merging~\cite{tomebolya2023token}} &   ViT-S/16$_{(r=2)}$ & 22 & 4.31 & 172.7 & 74.0 \\
        &  ViT-S/16$_{(r=4)}$  & 22  & 4.02  & 161.6 & 73.8  \\
        &  ViT-S/16$_{(r=8)}$  & 22  & 3.41  & 138.2 & 73.1  \\
        &  ViT-S/16$_{(r=10)}$ & 22  & 3.14  & 125.9 & 72.5  \\
        &  ViT-S/16$_{(r=12)}$ & 22  & 2.85  & 115.2 & 71.6  \\
        &  ViT-S/16$_{(r=14)}$ & 22  & 2.57  & 103.3 & 70.4  \\
        &  ViT-S/16$_{(r=16)}$ & 22  & 2.30  & 94.0  & 68.1  \\
        &  ViT-B/16$_{(r=2)}$  & 86  & 16.46 & 551.1 & 81.0  \\
        &  ViT-B/16$_{(r=4)}$  & 86  & 15.34 & 515.7 & 80.9  \\
        &  ViT-B/16$_{(r=8)}$  & 86  & 13.12 & 440.6 & 80.4  \\
        &  ViT-B/16$_{(r=10)}$ & 86  & 12.02 & 402.0 & 80.1  \\
        &  ViT-B/16$_{(r=12)}$ & 86  & 10.93 & 367.0 & 79.6  \\
        &  ViT-B/16$_{(r=14)}$ & 86  &  9.84 & 330.2 & 78.9  \\
        &  ViT-B/16$_{(r=16)}$ & 86  &  8.78 & 296.4 & 77.6  \\
        \midrule
        \multirow{3}{*}{Ours} & ViTTM-S$_{(64,64)}$    & 33 &  1.84 & 77.7 & 79.2  \\
        & ViTTM-B$_{(64,64)}$    & 127 &  7.08 & 234.1 & 82.9  \\
        & ViTTM-B$_{(49,196)}$ & 125  &  7.10 &  251.5 & 80.9 \\
        \bottomrule
    \end{tabular}
    \caption{
    Comparison of ViTTM with state-of-the-art methods on image classification (ImageNet-1K).
    Latency was measured on a 80GB A100 with batch size 256.
    \textit{Notes}:
    The ViT baseline model is the 224 resolution fine-tuned model from~\cite{dosovitskiy_image_2020}, available from \texttt{timm}~\cite{huggingface_timm}.
    LookupViT does not have a public implementation, as such we implement a version following the paper.
    Token Merging~\cite{tomebolya2023token} is applied to ViT-S/16 and ViT-B/16 models at various pruning rates ($r$) \textit{without} fine-tuning.
    }
    \label{tab:app:extra_1k}
\end{table*}